\documentclass{article}

\usepackage[dblblindworkshop, final]{neurips_2025}

\usepackage[utf8]{inputenc} 
\usepackage[T1]{fontenc}    
\usepackage{hyperref}       
\usepackage{url}            
\usepackage{booktabs}       
\usepackage{amsfonts}       
\usepackage{nicefrac}       
\usepackage{microtype}      
\usepackage{xcolor}         

\usepackage{graphicx}
\usepackage{amsmath}
\usepackage{tcolorbox}
\usepackage{fancyvrb}
\usepackage{float}
\tcbuselibrary{breakable}
\usepackage{fvextra}

\newtcolorbox{promptbox}{
    colback=blue!5!white,
    colframe=blue!75!black,
    boxrule=1pt,
    arc=3pt,
    left=3pt,
    right=3pt,
    top=3pt,
    bottom=3pt,
    fonttitle=\bfseries,
    breakable
}

\title{Layer Importance for Mathematical Reasoning is Forged in Pre-Training and Invariant after Post-Training}

\author{
\textbf{Aadim Nepal}$^{1*}$ \quad
\textbf{Safal Shrestha}$^{1}$ \quad
\textbf{Anubhav Shrestha}$^{1}$ \quad
\textbf{Minwu Kim}$^{1}$ \quad
\textbf{Jalal Naghiyev}$^{2}$ \\
\textbf{Ravid Shwartz-Ziv}$^{3}$ \quad
\textbf{Keith Ross}$^{1}$ \\
\\
$^{1}$ New York University Abu Dhabi \\
$^{2}$ Technical University of Munich \\
$^{3}$ NYU Center for Data Science \\
\\
}

\begin{document}

\maketitle

\begin{abstract}
Large language models improve at math after instruction tuning, reinforcement learning, or knowledge distillation. We ask whether these gains come from major changes in the transformer layers or from smaller adjustments that keep the original structure. Using layer-wise ablation on base and trained variants, we find that math reasoning depends on a few critical layers, which stay important across all post-training methods. Removing these layers reduces math accuracy by as much as 80\%, whereas factual recall tasks only show relatively smaller drops. This suggests that specialized layers for mathematical tasks form during pre-training and remain stable afterward. As measured by Normalized Mutual Information (NMI), we find that near these critical layers, tokens drift from their original syntactic clusters toward representations aligned with tokens less syntactically related but potentially more useful for downstream task.
\footnotetext[1]{Code available at: \url{https://github.com/anonymous-xyz-anonymous/Ablation}.}
\footnotetext[2]{*Correspondence to: \texttt{aadim.nepal@nyu.edu}.}

\end{abstract}

\section{Introduction}

The capabilities of large language models (LLMs) have been improved through different post-training methods such as instruction tuning and reinforcement learning with human feedback, knowledge distillation from a teacher model, and reinforcement learning with verifiable rewards (RLVR) \citep{ouyang2022training, guo2025deepseek, lambert2024tulu3, bai2023qwen, openai2025o1}. Yet, the following question remains unexplored:

\textit{Do post-training methods fundamentally change how models process mathematical problems or just make small adjustments to existing structures?}

In this paper, we investigate this question through layer ablation experiments on two model families, Llama-3.1-8B \citep{meta2024llama3_1} and Qwen-2.5-7B \citep{qwen2.5_2024}, examining four variants for each: the pre-trained base model, an instruction-tuned model, a knowledge-distilled model, and an RLVR-trained model. We evaluate these models on two mathematical reasoning benchmarks, GSM8K \citep{cobbe2021training} and MATH500 \citep{hendrycks2021measuring}. We find that post-training largely preserves the layer importance structure in mathematical reasoning: the set of critical layers - whose removal leads to substantial performance drop - remains largely invariant across all four model variants.
Additionally, we conducted the same experiment on the TriviaQA \citep{joshi2017triviaqa} factual recall task. Here, we observed a different layer importance pattern: no specific critical layers were found. Instead, removing any layer results in a smaller, more consistent drop in performance. Previous work \citep{meng2022locating, NEURIPS2019_2c601ad9, voita2019analyzingmultiheadselfattentionspecialized, elhage2021mathematical, pochinkov2024investigatingneuronablationattention} has mainly analyzed individual components of transformers—such as neurons, attention heads—but has not examined how post-training affects entire layer structures, or compared these effects between reasoning and non-reasoning  \citep{lad2025robustness, sun2024painter, zhang2024investigating, yu2024superweight}. 

To investigate what happens at these critical layers, we apply a Normalized Mutual Information (NMI) analysis to track how token representations evolve across layers. Tokens are first clustered into families based on their initial semantics from the first layer. As the model approaches the critical layers, an increasing number of tokens drift across clusters, indicating that originally unrelated tokens begin to form new relationships at these points.

\begin{figure}[t]
\centering
\includegraphics[width=0.9\textwidth]{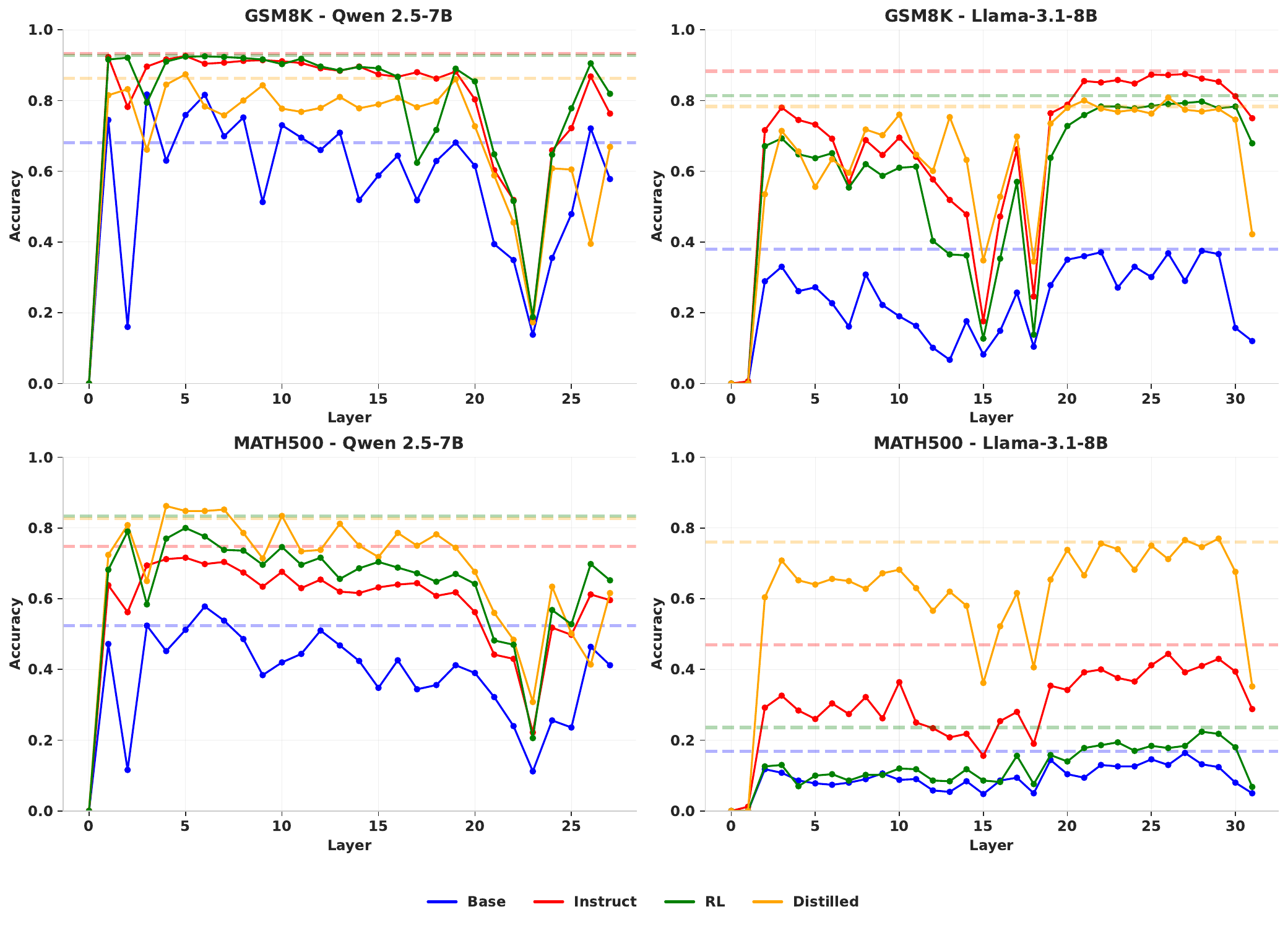}
\caption{The plots show model accuracy (Y-axis) on GSM8K and MATH500 when a single transformer layer (X-axis) is zeroed out. The performance of all model variants drops substantially when specific layers are removed (layer 23 for Qwen, layers 15 and 18 for Llama), a pattern that remains consistent across different datasets and post-training methods. Dashed lines indicate the original, un-ablated performance.} 
\label{fig:math_benchmarks}
\end{figure}

\section{Related Work}
\label{gen_inst}
\paragraph{Critical Layers and Ablation.}
Transformers respond unevenly to structural changes: some layers can be removed or reordered with little effect, while others are essential, especially for reasoning tasks \citep{sun2024painter}. Layer ablation has revealed “cornerstone layers,” whose removal collapses performance \citep{zhang2024investigating}, and “super weights,” a small parameter set critical for fluency \citep{yu2024superweight}. Zero-ablation studies further show that uncertainty and factuality share circuits \citep{teplica-etal-2025-sciurus}. Earlier work on BERT found layer specialization resembling an NLP pipeline \citep{tenney2019bert, vanaken2019bertqa}. Other studies identified four inference stages across depth using ablation and probing \citep{lad2025robustness}, and many focus on smaller components—heads, neurons, or MLP blocks \citep{meng2022locating, NEURIPS2019_2c601ad9, voita2019analyzingmultiheadselfattentionspecialized, pochinkov2024investigatingneuronablationattention, elhage2021mathematical}. While most rely on zero-ablation, we use systematic layer-wise ablation to show that pre-training establishes critical layers for reasoning, and we extend this to study post-training effects. This complements findings that post-training preserves knowledge and truthfulness \citep{du2025posttrainingreshapesllmsmechanistic}, but does not address layer-level stability across reasoning vs. non-reasoning tasks.

\paragraph{Representational Analysis with NMI.}
Prior work has analyzed internal representations to explain why certain layers matter, often using probing tasks or attention analysis \citep{lad2025robustness,tigges2023linear,marks2023geometry}. While these approaches can reveal specific mechanisms, they require choosing properties to probe in advance and are less suited to generative tasks. Normalized Mutual Information (NMI), by contrast, compares clustering structures across layers \citep{strehl02a} and has shown, for example, that BERT encodes a linguistic hierarchy \citep{jawahar2019what} and that representations shift during QA \citep{vanaken2019bertqa}. Intermediate layers in particular appear to hold rich, geometrically structured features \citep{skean2025layer, gurnee2023language}. Our work uses NMI differently: instead of probing for specific linguistic features, we measure the extent of divergence from the input layer’s clustering to highlight where semantic relationships form.

\begin{figure}[t]
\centering
\includegraphics[width=0.9\textwidth]{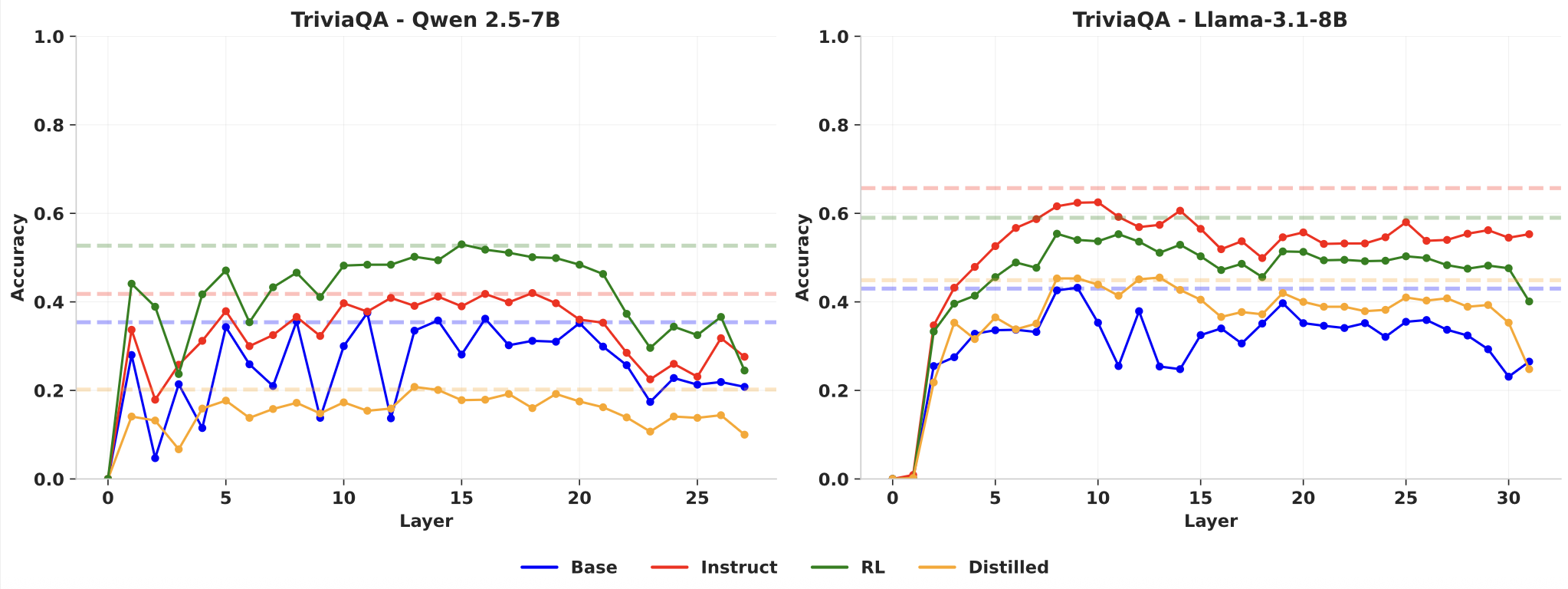}
\caption{Layer ablation results on the TriviaQA factual recall task. The left plot shows performance for Qwen 2.5-7B models, and the right plot shows performance for Llama 3.1-8B models when individual layers are zeroed out. The X-axis represents the layer index (0-32), and the Y-axis shows the accuracy.}
\label{fig:non_math_tasks}
\end{figure}

\section{Experiments and Results}
\label{results}

We perform systematic layer-wise zero ablation experiments to assess the importance of layers in the model. We choose the Qwen 2.5-7B and Llama-3.1-8B models as the primary base models for analysis. We take their respective post-trained variants as well for analysis. For math reasoning, we choose GSM8K \citep{cobbe2021training} and MATH \citep{hendrycks2021measuring} and for factual recall, we choose TriviaQA \cite{joshi2017triviaqa}. Unlike past studies which primarily focus on classification benchmarks \citep{chen2024streamlining,lad2025robustness}, we opt for generative tasks for both math and non-math. This removes chances of random guesses and provides more accurate estimate of model abilities.

Furthermore, we track representational shifts across layers using Normalized Mutual Information (NMI), which measures the similarity between token clusterings at different layers. NMI ranges from 0 (independent) to 1 (identical), providing a simple way to quantify how far representations drift from their original input-layer clusters (Full details on the methodology including model details, ablation, and NMI are provided in Appendix \ref{detailed_methodology}).

\subsection{Ablation studies reveal stable critical layers}
Our ablation studies reveal that mathematical reasoning in LLMs depends on a small set of stable critical layers. In Figure \ref{fig:math_benchmarks}, for Qwen, performance drops sharply when ablating layer 23, while Llama is most sensitive at layers 15 and 18. Removing these layers still provides coherent responses but we find the model makes simple airthmetic mistakes when these layers are removed (\autoref{fig:arith_logic}). Notably, these critical layers remain consistent after post-training: instruction tuning, distillation, and RLVR do not shift which layers are essential. We also find vulnerabilities at boundary layers: early layers are crucial for coherent language, consistent with “super-weight” findings \citep{yu2024superweight}. For Qwen, ablating the first layer breaks fluency, while for Llama, the first two layers are critical. Some final layers in Llama also matter, reflecting architectural differences between models. Importantly, mathematical reasoning is far more sensitive to layer ablation than factual recall: removing critical layers causes accuracy collapses of 60–80\%, whereas factual recall shows only modest, distributed declines of 10–30\%. Interestingly, (as shared in Appendix~\ref{qwen_twins}), Qwen’s layer importance structures are similar across different model sizes as well, which could be because of their potentially shared pretraining schemes.

We observed a large drop at layer 2 of the Qwen base model. On closer inspection, most responses after removing layer 2 were incoherent. In contrast, removing layer 23 still produced coherent responses (\autoref{sec:sample-responses}), but with many arithmetic errors, as shown in \autoref{fig:arith_logic}. Finally, the plots in Figure 1 were averaged over hundreds of runs, and we confirm that the standard deviation at each point is minimal, so the results are stable and not due to random noise




\subsection{Clustering Analysis}

\begin{figure}[t]
\centering
\includegraphics[width=0.8\linewidth]{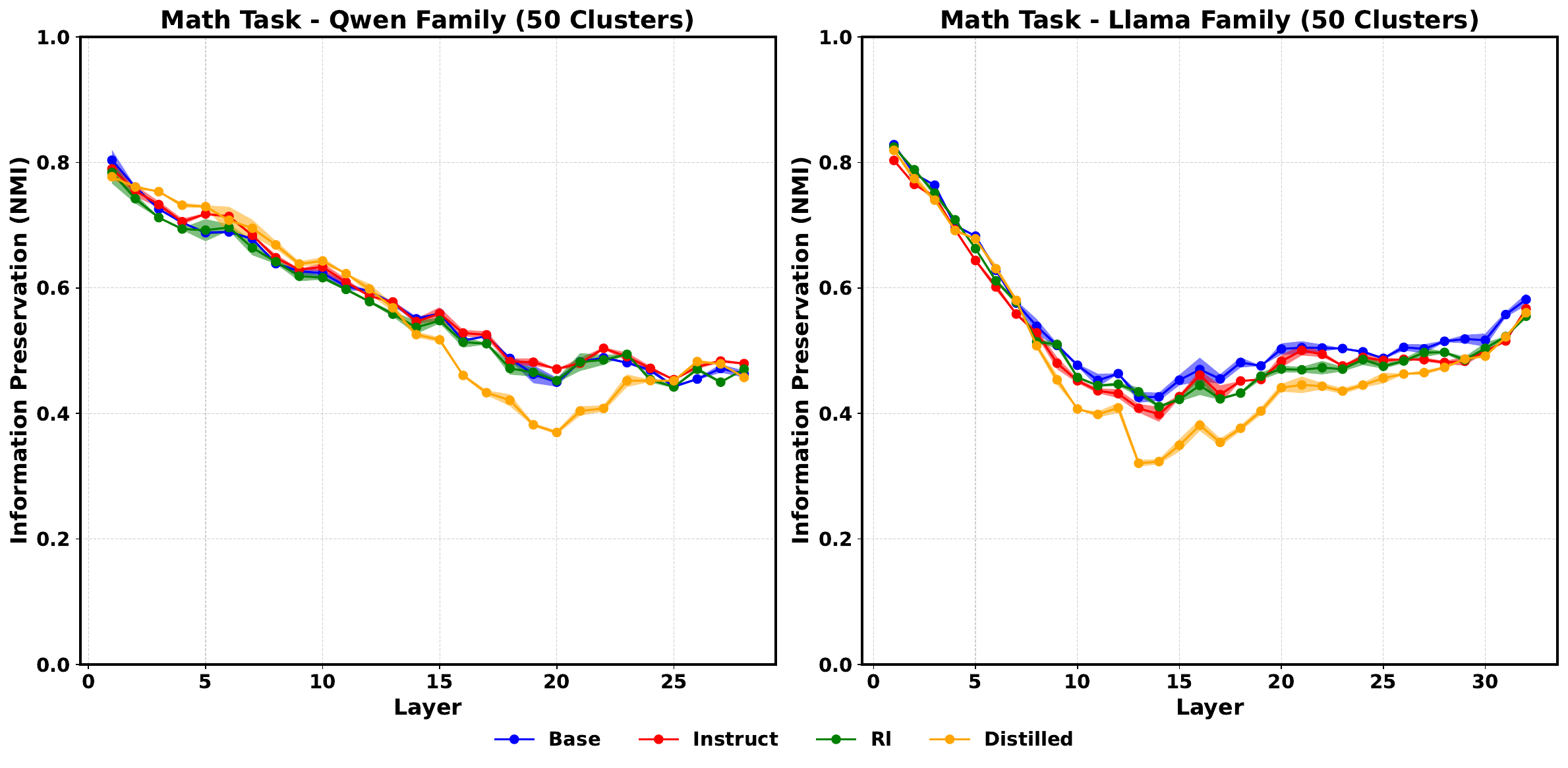}
\caption{The plots show the NMI score (Y-axis) at each transformer layer (X-axis), calculated relative to the token clusters at Layer 0. The observed trends are robust to the number of clusters (k) used for the analysis, with similar results for k-values between 10 and 70. The choice of 50 here is arbitrary. Shaded region denotes standard deviation over 5 runs. For each run and each model, the 20 problems were selected randomly, so we are looking at over 100 math problems over 8 model families, which is about 800 problems in total.}
\label{fig:nmi}
\end{figure}

To gain some preliminary insight into the role of the critical layers, we performed a token clustering analysis using the procedure described in Appendix \ref{sec:nmi}. For each task, we constructed a single large-context prompt. For mathematical reasoning, the prompt concatenated 20 random problems and solutions from MATH500. Details are provided in Appendix~\ref{sec:nmi-prompts}.

Figure~\ref{fig:nmi} shows that for math tasks, the NMI score starts high and then drops in the middle-to-late layers, forming an “elbow” region. For Qwen models this appears between layers 20–25, and for Llama between layers 13–18. We use Layer 0 as the baseline because prior work shows that early layers tend to group tokens by syntactic type \citep{vanaken2019bertqa, jawahar2019what}. A lower NMI score means that token clusters differ more from this baseline, indicating more mixing between token types. The layers identified as critical in our ablation experiments fall within this elbow region, suggesting a possible connection between token clustering changes and layer importance. In contrast, TriviaQA shows little change in NMI across layers (\autoref{fig:trivia_nmi}). Token clusters remain close to the Layer 0 baseline, likely reflecting the lower token diversity in these prompts compared to math problems. This matches our ablation results, where no single critical layer was found for factual recall.


\subsection{Qualitative Cluster Analysis of Token Representations}
\label{app:qualitative-clusters}

To complement our NMI analysis, we performed $k$-means clustering
($k=10$) on hidden states at each layer for a representative mathematical reasoning prompt.
This qualitative approach offers concrete illustrations of how token representations evolve
through the network and helps interpret the quantitative trends captured by the NMI score.

\subsubsection{Layer 0: Surface-Form Grouping}
At the embedding layer (Layer~0), clusters are dominated by raw orthographic units and
formatting tokens:
\begin{itemize}
    \item \textbf{Symbols:} repeated tokens such as ``\verb|\frac|'', ``$=$'', ``$-$'', and braces ``\{ \{ \{'' form distinct clusters.
    \item \textbf{Function words:} tokens such as ``the'', ``of'', ``and'', ``by'' co-cluster.
    \item \textbf{Special tokens:} \verb|<|im\_start|>|, \texttt{system}, \texttt{user}, and the initial prompt words appear together in a large mixed cluster.
\end{itemize}
This behavior reflects what NMI quantifies as high similarity to the Layer~0 baseline: tokens
are grouped by surface type, with little semantic integration.

\subsubsection{Layer 23: Emergence of Semantic Roles}
By Layer~23 (the empirically identified critical layer for Qwen), clusters reflect semantically
meaningful groupings aligned with the problem-solving process:
\begin{itemize}
    \item \textbf{Problem setup (Cluster 3):} tokens representing the given equations, e.g.\ ``$2x=3y=-z$'' and ``$6x=-y=-4z$''.
    \item \textbf{Parametric reformulation (Cluster 4):} ``parametric'', ``form'', ``direction vector'', and ``line'' co-occur.
    \item \textbf{Mathematical operations (Clusters 2 and 5):} fractions such as ``$\tfrac{k}{2}$'', ``$\tfrac{1}{3}$'', and symbols like ``$=$'' appear in equation-level context.
    \item \textbf{Conclusion (Cluster 7):} ``angle'', ``dot product'', ``orthogonal'', and ``90'' cluster together, corresponding to the final reasoning step.
\end{itemize}

\subsubsection{Relation to NMI}
The NMI analysis measures \emph{how much} cluster structures diverge
from Layer~0. The qualitative results here illustrate \emph{what} those divergences look like.
At critical layers such as Layer~23 in Qwen, where NMI reaches a local minimum, we see
that tokens reorganize into semantically interpretable clusters that align with reasoning
steps. This convergence of quantitative (NMI) and qualitative (clustering) evidence supports
the interpretation that critical layers are precisely those where surface-level groupings are
broken apart and recomposed into task-specific structures. In summary, from this qualitative token analysis, we observe that at early layers, tokens group mainly by syntactic meaning, for example brackets clustering with other brackets. Around the critical layers, these clusters reorganize into mixed groups that combine brackets with numbers or symbols. This supports the interpretation that tokens are drifting away from syntactic meaning and forming meaningful semantic relationships.

\section{Limitations}


We test on Qwen and Llama models at the 7B–8B scale, though broader model sizes and families should be explored for generalizability. We use NMI because it provides a high-level view of how token interactions evolve across layers, however, it cannot definitively establish reasoning. Also, other techniques like probing is better suited for classification tasks and requires deciding in advance what to probe for, while attention analysis typically focuses on individual prompts or tokens and thus fails to capture overall representational shifts \citep{lad2025robustness,tigges2023linear,marks2023geometry}. For completeness, we also evaluated layer importance through residual contributions (Appendix~\ref{residual}), but found no strong correlation with critical layers. This potentially suggests that hidden-state–based methods may not reliably identify task-relevant layers in generative mathematical reasoning, though further evidence is needed.

\section{Future Work and Conclusion}

We show that mathematical reasoning depends on a small set of critical layers that remain stable after post-training, whereas this behavior does not hold for non-reasoning tasks such as factual recall. Using additional token-clustering analysis, we provide an initial high-level explanation for why these layers may be particularly important for reasoning.
Future work could explore targeted fine-tuning strategies; for instance, freezing all other layers and updating only these critical ones to improve mathematical reasoning efficiency. In addition, a more fine-grained mechanistic-interpretability analysis on these layers may shed light on why they are critical at a more granular level.


\bibliographystyle{unsrt}
\bibliography{references}







\appendix

\section{Appendix}
\subsection{Acknowledgement}
This work is submitted in part by the NYU Abu Dhabi Center for Artificial Intelligence and
Robotics, funded by Tamkeen under the Research Institute Award CG010. Some experiments were
carried out on the High Performance Computing resources at New York University Abu Dhabi.

\subsection{Experimental Setup}
\textbf{Model Configuration.} All models used consistent sampling parameters: temperature=0.7, top-p=0.9, max tokens=8000. Models were loaded using vLLM  \citep{kwon2023efficient} with 0.9 GPU memory utilization and eager execution for reproducibility.

\textbf{Datasets.} We evaluated on three datasets: GSM8K (1000 randomly sampled grade school math problems), MATH500 (500 problems across seven mathematical categories and five difficulty levels), and TriviaQA (1000 randomly sampled factual recall questions using RC no-context configuration).

\textbf{Hardware.} Experiments ran on a single NVIDIA A100 80GB GPU with 128GB system memory. Each complete layer ablation analysis required 4-11 hours depending on dataset size and model variant.

\subsection{Detailed Methodology}
\label{detailed_methodology}

\begin{figure}[H]
    \centering
    \includegraphics[width=0.6\textwidth]{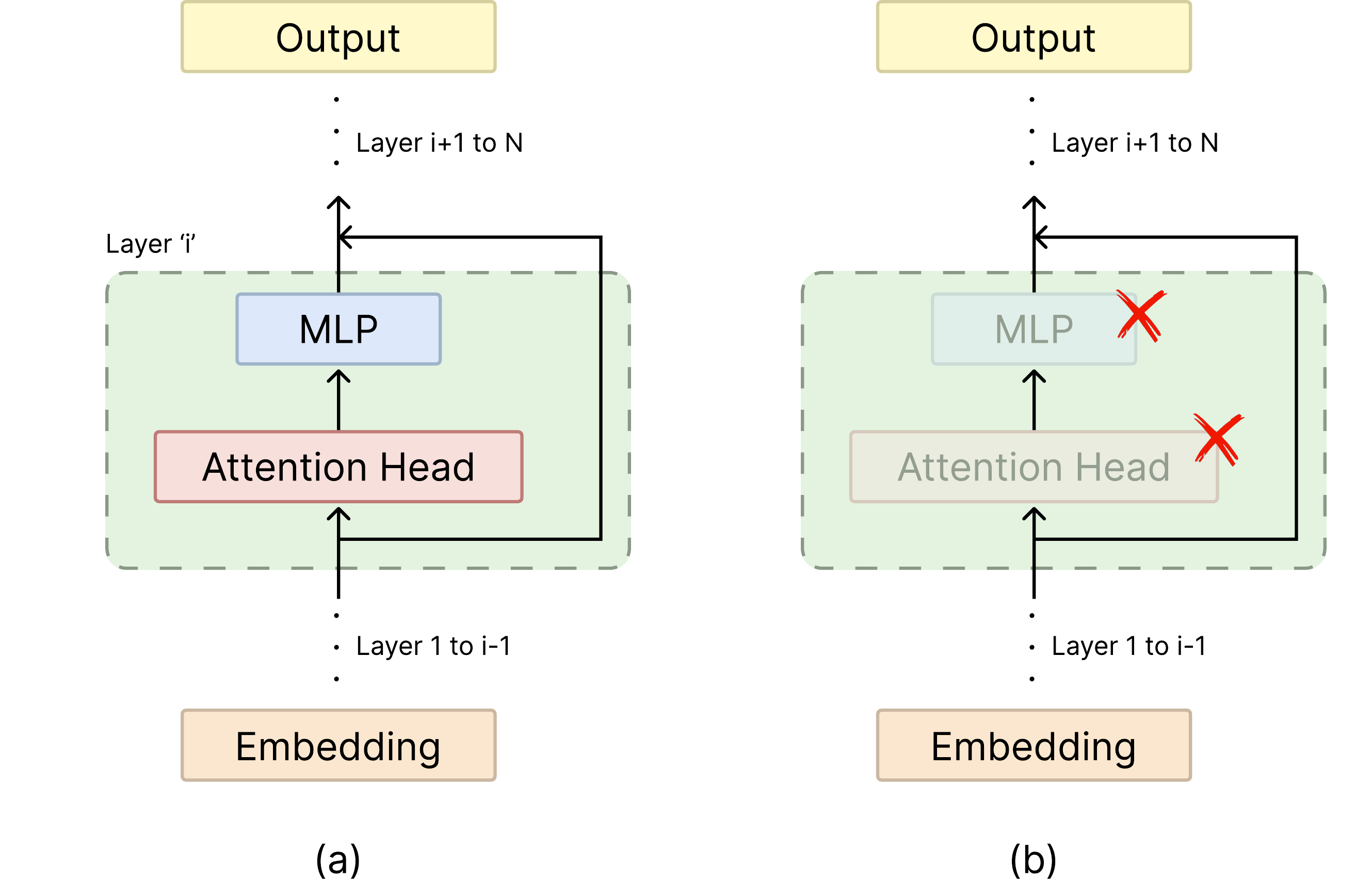}
    \caption{Zero ablation technique illustration. (a) Normal transformer layer with active MLP and attention. (b) Ablated layer with MLP and attention parameters set to zero, effectively nullified due to the skip connection}
    \label{fig:methodology}
\end{figure}

\subsubsection{Zero-Ablation of Transformer Layers}

As illustrated in Figure \ref{fig:methodology}, our zero-ablation procedure systematically nullifies all parameters within a target layer while preserving the overall model architecture through its residual connections. For each target layer $\ell$ in a given model, we set the weight matrices and bias vectors in both the multi-head attention and MLP sublayers to zero. Specifically, for the multi-head attention mechanism, we ablate:
\begin{align*}
W_Q^{(\ell)}, W_K^{(\ell)}, W_V^{(\ell)}, W_O^{(\ell)} &= \mathbf{0} \\
b_Q^{(\ell)}, b_K^{(\ell)}, b_V^{(\ell)}, b_O^{(\ell)} &= \mathbf{0}
\end{align*}
For the MLP component, we ablate:
\begin{align*}
W_{\text{gate}}^{(\ell)}, W_{\text{up}}^{(\ell)}, W_{\text{down}}^{(\ell)} &= \mathbf{0} \\
b_{\text{gate}}^{(\ell)}, b_{\text{up}}^{(\ell)}, b_{\text{down}}^{(\ell)} &= \mathbf{0}
\end{align*}
Here, $W_Q, W_K, W_V$ are the query, key, and value weight matrices; $W_O$ is the output projection; and $W_{\text{gate}}, W_{\text{up}}, W_{\text{down}}$ are the gating, up-projection, and down-projection matrices in the MLP.

When the self-attention and MLP sub-layers are zeroed out, the operation of layer $\ell$ is reduced to its residual connections and normalization steps. Since we are looking at a pre-norm architecture, the layer becomes an identity layer:
$$
\mathbf{h}^{(\ell)} = \mathbf{h}^{(\ell-1)}
$$


We conduct this analysis on two model families, Llama-3.1-8B \citep{meta2024llama3_1} and Qwen-2.5-7B \citep{qwen2.5_2024}, examining four variants for each:
(1) the pre-trained base models,
(2) instruction-tuned models: Llama-3.1-8B-Instruct \citep{meta2024llama3_1} and Qwen-2.5-7B-Instruct \citep{qwen2.5_2024},
(3) knowledge-distilled models: DeepSeek-R1-Distill-Llama-8B and DeepSeek-R1-Distill-Qwen-7B \citep{guo2025deepseek}, and
(4) RLVR-trained models: Llama-3.1-8B-SimpleRL-Zoo \citep{zeng2025simplerlzooinvestigatingtamingzero} and Open-Reasoner-Zero-7B \citep{hu2025openreasonerzeroopensourceapproach}.
Our evaluation spans mathematical reasoning (GSM8K \citep{cobbe2021training}, MATH500 \citep{hendrycks2021measuring}) and factual recall (TriviaQA \citep{joshi2017triviaqa}) to distinguish task-specific processing patterns.

\subsubsection{Representational Analysis via Normalized Mutual Information (NMI)} \label{sec:nmi}

To understand \textit{why} certain layers are critical, we analyze the evolution of the model's internal representations using Normalized Mutual Information (NMI) \citep{strehl02a}. While prior work has qualitatively described how token representations cluster across transformer layers \citep{vanaken2019bertqa}, our approach introduces a quantitative measure. We compute the NMI between the clustering of hidden states at layer 0 and the clustering of hidden states at each subsequent layer $\ell$. This enables us to track how far each layer's cluster structure has diverged from the initial structure. See Appendix~\ref{sec:nmi-prompts} for more details on the prompts used.

Our procedure is as follows:
\begin{enumerate}
    \item \textbf{Extract Hidden States:} For a given task input, we perform a single forward pass and store the hidden state representations for each layer. For each layer $\ell$, we have a sequence of $T$ token representations, $\mathbf{h}^{(\ell)} = (\mathbf{h}^{(\ell)}_1, \dots, \mathbf{h}^{(\ell)}_T)$, where each $\mathbf{h}^{(\ell)}_t \in \mathbb{R}^{d}$ and $d$ is the dimension of the hidden state.

    \item \textbf{Establish Baseline Clustering:} We apply K-Means clustering to the $T$ vectors from layer 0, $(\mathbf{h}^{(0)}_1, \dots, \mathbf{h}^{(0)}_T)$, to establish a set of baseline clusters, $\mathbf{C}^{(0)} = \{\mathbf{C}^{(0)}_1, \dots, \mathbf{C}^{(0)}_K\}$. Note that we use zero-indexing, so layer 0 refers to the output of the first transformer layer.

    \item \textbf{Cluster Subsequent Layers:} For each subsequent layer $\ell > 0$, we apply K-Means with the same number of clusters, $K$, to its hidden state vectors $(\mathbf{h}^{(\ell)}_1, \dots, \mathbf{h}^{(\ell)}_T)$ to obtain a new set of clusters, $\mathbf{C}^{(\ell)} = \{\mathbf{C}^{(\ell)}_1, \dots, \mathbf{C}^{(\ell)}_K\}$.

    \item \textbf{Calculate NMI:} We measure the similarity between the baseline clustering $\mathbf{C}^{(0)}$ and each subsequent layer's clustering $\mathbf{C}^{(\ell)}$. Given two clusterings $\mathbf{C}$ and $\mathbf{D}$, the NMI is formulated using the arithmetic mean for normalization: \citep{strehl02a}
    $$
    \text{NMI}(\mathbf{C}, \mathbf{D}) = \frac{I(\mathbf{C}, \mathbf{D})}{(H(\mathbf{C}) + H(\mathbf{D}))/2}
    $$
    This corresponds to the default method in the \texttt{sklearn.metrics.normalized\_mutual\_info\_score} function used in our analysis. A detailed breakdown of each term—Mutual Information ($I$) and Entropy ($H$)—is provided in Appendix~\ref{sec:detailed_nmi}.
\end{enumerate}

\subsubsection{Detailed NMI Calculation} \label{sec:detailed_nmi}

This section provides a detailed breakdown of the terms used to calculate the Normalized Mutual Information (NMI) score, as referenced in \autoref{sec:nmi}. The calculation follows the default implementation in the \texttt{scikit-learn} package. The final formula uses arithmetic mean normalization:
$$
\text{NMI}(C, D) = \frac{I(C, D)}{(H(C) + H(D)) / 2}
$$
The calculation begins by defining the properties of the clusterings. Let $C = \{c_1, \dots, c_{|C|}\}$ and $D = \{d_1, \dots, d_{|D|}\}$ be two different clusterings of the same $N$ tokens. The probability of a token belonging to a specific cluster $c_i \in C$ is $P(i) = \frac{|c_i|}{N}$, and for a cluster $d_j \in D$ it is $P(j) = \frac{|d_j|}{N}$. The joint probability of a token belonging to both cluster $c_i$ and cluster $d_j$ is $P(i, j) = \frac{|c_i \cap d_j|}{N}$.

Using these probabilities, we first calculate the entropy of each clustering independently. The entropy $H(C)$ measures the uncertainty or disorder of a single clustering and is defined as:
$$
H(C) = - \sum_{i=1}^{|C|} P(i) \log(P(i))
$$
The entropy $H(D)$ is calculated identically for the second clustering.

Next, the mutual information, $I(C, D)$, measures the information shared between the two clusterings. It quantifies the reduction in uncertainty about one clustering given knowledge of the other. It is defined as:
$$
I(C, D) = \sum_{i=1}^{|C|} \sum_{j=1}^{|D|} P(i, j) \log\left(\frac{P(i, j)}{P(i)P(j)}\right)
$$
These components are then combined in the final formula to produce the normalized score, which provides a robust measure of similarity between cluster structures on a scale from 0 to 1. Note that the \texttt{scikit-learn} implementation uses the natural logarithm.

\subsection{Prompt Templates} \label{sec:prompt-templates}

We used model-specific prompts optimized for each architecture family.

\textbf{Mathematical Reasoning (GSM8K \& Math500):}

\noindent\textit{Qwen Models:}
\begin{promptbox}
\begin{Verbatim}[fontsize=\small]
<|im_start|>system
Please reason step by step, and put your final answer within \boxed{}.<|im_end|>
<|im_start|>user
{problem}
<|im_end|>
<|im_start|>assistant
\end{Verbatim}
\end{promptbox}

\noindent\textit{Llama Base:}
\begin{promptbox}
\begin{Verbatim}[fontsize=\small]
Question: {problem}
Answer: Let's think step by step.
\end{Verbatim}
\end{promptbox}

\noindent\textit{Llama Instruct:}
\begin{promptbox}
\begin{Verbatim}[fontsize=\small]
<|begin_of_text|><|start_header_id|>system<|end_header_id|>
You are a helpful mathematics assistant. Please solve the problem step by 
step and provide your final answer within \boxed{}.<|eot_id|>
<|start_header_id|>user<|end_header_id|>
{problem}<|eot_id|>
<|start_header_id|>assistant<|end_header_id|>
\end{Verbatim}
\end{promptbox}

\noindent\textit{DeepSeek Distilled Models:}
\begin{promptbox}
\begin{Verbatim}[fontsize=\small]
A conversation between User and Assistant. The User asks a question, and the 
Assistant solves it. The Assistant first thinks about the reasoning process 
in the mind and then provides the User with the answer. The reasoning process 
is enclosed within <think> </think> and answer is enclosed within 
<answer> </answer> tags, respectively.
User: {problem}
Assistant: <think>
\end{Verbatim}
\end{promptbox}

\noindent\textit{Llama RL Model:}
\begin{promptbox}
\begin{Verbatim}[fontsize=\small]
<|im_start|>system
You are a helpful assistant.<|im_end|>
<|im_start|>user
{problem}
Please reason step by step, and put your final answer within \boxed{}.<|im_end|>
<|im_start|>assistant
\end{Verbatim}
\end{promptbox}

\textbf{Factual Recall (TriviaQA):}

\noindent\textit{Qwen Models:}
\begin{promptbox}
\begin{Verbatim}[fontsize=\small]
<|im_start|>system
Please answer the trivia question directly and concisely.<|im_end|>
<|im_start|>user
{question}
<|im_end|>
<|im_start|>assistant
\end{Verbatim}
\end{promptbox}

\noindent\textit{Llama Base \& RL:}
\begin{promptbox}
\begin{Verbatim}[fontsize=\small]
Question: {question}
Answer: Let's think step by step.
\end{Verbatim}
\end{promptbox}

\noindent\textit{Llama Instruct:}
\begin{promptbox}
\begin{Verbatim}[fontsize=\small]
<|begin_of_text|><|start_header_id|>system<|end_header_id|>
You are a helpful assistant. Please answer the trivia question directly 
and accurately.<|eot_id|>
<|start_header_id|>user<|end_header_id|>
{question}<|eot_id|>
<|start_header_id|>assistant<|end_header_id|>
\end{Verbatim}
\end{promptbox}

\noindent\textit{DeepSeek Distilled Models:} 
\begin{promptbox}
\begin{Verbatim}[fontsize=\small]
A conversation between User and Assistant. The User asks a question, and the 
Assistant solves it. The Assistant first thinks about the reasoning process 
in the mind and then provides the User with the answer. The reasoning process 
is enclosed within <think> </think> and answer is enclosed within 
<answer> </answer> tags, respectively.
User: {question}
Assistant: <think>
\end{Verbatim}
\end{promptbox}

\subsection{Answer Extraction and Evaluation}
\textbf{Mathematical Problems.} We used the \texttt{math\_verify} library to parse numerical answers from model outputs, handling various formats including \texttt{\textbackslash boxed\{\}} notation, explicit numerical statements, and embedded calculations.

\textbf{Factual Recall.} For TriviaQA, we employed case-insensitive substring matching, checking for ground truth answers (including aliases) within model responses.

\subsection{Sample Responses by Important Layers:}\label{sec:sample-responses}

Below are the sample responses of base model and layer ablated models. Specifically, we show the responses for the original base model and the model with most significant layer ablated.

\noindent\textit{Sample Question:}
\begin{promptbox}
\begin{Verbatim}[fontsize=\small, breaklines=true]
Joy can read 8 pages of a book in 20 minutes. How many hours will it take her to read 
120 pages?
\end{Verbatim}
\end{promptbox}

\noindent\textit{Qwen Base Original (Correct):}

\begin{promptbox}
\begin{Verbatim}[fontsize=\small, breaklines=true]
Joy can read 8 pages in 20 minutes, so she can read 8/20 = 0.4 pages per minute.
To read 120 pages, it will take her 120/0.4 = 300 minutes.
There are 60 minutes in an hour, so it will take her 300/60 = 5 hours.
The answer is 5.
\end{Verbatim}
\end{promptbox}

\noindent\textit{Qwen Base Layer 23 Ablated (Incorrect):}

\begin{promptbox}
\begin{Verbatim}[fontsize=\small, breaklines=true]
To determine how many hours it will take Joy to read 120 pages, we can follow these
steps:
1. Find the number of pages Joy reads per minute: Joy reads 8 pages in 20 minutes, so 
she reads 8/20 = 0.4 pages per minute.
2. Determine the number of minutes it takes to read 120 pages: Since Joy reads 0.4 
pages per minute, it will take her 120/0.4 = 2400 minutes to read 120 pages.
3. Convert the minutes to hours: Since 1 hour is equal to 60 minutes, we can convert 
2400 minutes to hours by dividing by 60: 2400/60 = 120/3 = 60/2 = 30 hours.

So, it will take Joy 30 hours to read 120 pages.
\end{Verbatim}
\end{promptbox}


\noindent\textit{Llama Base Original (Correct):}
\begin{promptbox}
\begin{Verbatim}[fontsize=\small, breaklines=true]
Joy can read 8 pages in 20 minutes. So, she can read 1 page in 20/8 minutes. That is,
she can read 1 page in 2.5 minutes. So, she can read 120 pages in 120 x 2.5 minutes. 
That is, she can read 120 pages in 300 minutes. That is, she can read 120 pages in 5
hours.
\end{Verbatim}
\end{promptbox}

\noindent\textit{Llama Base Layer 15 Ablated (Incorrect):}

\begin{promptbox}
\begin{Verbatim}[fontsize=\small, breaklines=true]
Joy can read 8 pages in 20 minutes. So, she can read 1 page in 20/8 minutes. That 
means she can read 1 page in 4 minutes. Now, we know that 1 page takes 4 minutes. 
So, 120 pages will take 120 x 4 = 480 minutes. 480 minutes is 480/60 = 96 hours. 
So, Joy will take 96 hours to read 120 pages.
\end{Verbatim}
\end{promptbox}

\subsection{NMI Prompts} \label{sec:nmi-prompts}

We perform NMI analysis using a single prompt that concatenates 20  math problems from the MATH500 dataset, each followed by its solution. Note that each of these 20 problems were sampled randomly for each run and for each model family. The NMI score were averaged across 5 runs. These problems were selected by taking two from each of the seven  categories in the dataset to ensure broad coverage. The prompt is used across all model variants and passed through the model once - without any decoding - to extract hidden states from each layer. The full procedure is described in \autoref{sec:nmi}. A sample excerpt showing 2 of one of the 20 sampled problems is shown below.\\

Similarly, we perform NMI analysis on the TriviaQA task using a single prompt that concatenates 20 random question-answer pairs. As above, the prompt is passed through the model once - without any decoding - to extract hidden states across layers. A sample excerpt showing 5 of the 20 questions is shown below.

Our main goal for this analysis was to see how the model organizes its representations when given a prompt with a wide variety of tokens and concepts. To achieve this diversity, we concatenated multiple question-and-solution pairs into a single prompt. We used a single forward pass on this long prompt because of the unique challenge it presents. While the self-attention mechanism allows every token to see every other token, the need to form a coherent representation for each distinct problem creates a strong incentive for the model to learn to differentiate between their contexts. This is a more demanding task than processing a simple, isolated prompt, and it gives us a better view of how the model uses its representations to manage and insulate different contexts. While ideally, we might use a single, extremely long problem, our approach is a practical way to achieve the high diversity needed for this analysis.

\clearpage  

\noindent\textit{Sample Prompt for NMI Analysis on MATH task:}

\noindent\begin{promptbox}
Question: Convert the point $(0,3)$ in rectangular coordinates to polar coordinates.  Enter your answer in the form $(r,\theta),$ where $r > 0$
and $0 \le \theta < 2 \pi.$ 

\vspace{6pt}

Answer: We have that $r = \sqrt{0^2 + 3^2} = 3.$  Also, if we draw the line connecting the origin and $(0,3),$ this
line makes an angle of $\frac{\pi}{2}$ with the positive $x$-axis. Therefore, the polar coordinates are $\boxed{\left( 3, \frac{\pi}{2} \right)}.$

\vspace{12pt}

Question: The set of points $(x,y,z)$ that satisfy \[2x = 3y = -z\] is a line. The set of points $(x,y,z)$ that satisfy \[6x = -y = -4z\] is another
line. Find the angle between these lines, in degrees. 

\vspace{6pt}

Answer: For the first line, let $t = 2x = 3y = -z.$  Then \[\begin{pmatrix} x \\ y \\ z
\end{pmatrix} = \begin{pmatrix} t/2 \\ t/3 \\ -t \end{pmatrix} = \frac{t}{6} \begin{pmatrix} 3 \\ 2 \\ -6 \end{pmatrix}.\] Thus, the direction vector of the
first line is $\begin{pmatrix} 3 \\ 2 \\ -6 \end{pmatrix}.$ For the second line, let $t = 6x = -y = -4z.$  Then \[\begin{pmatrix} x \\ y \\ z \end{pmatrix}
= \begin{pmatrix} t/6 \\ -t \\ -t/4 \end{pmatrix} = \frac{t}{12} \begin{pmatrix} 2 \\ -12 \\ -3 \end{pmatrix}.\] Thus, the direction vector of the second line
is $\begin{pmatrix} 2 \\ -12 \\ -3 \end{pmatrix}.$ Note that \[\begin{pmatrix} 3 \\ 2 \\ -6 \end{pmatrix} \cdot \begin{pmatrix} 2 \\ -12 \\ -3 \end{pmatrix} =
0.\] Hence, the angle between the lines is $\boxed{90^\circ}.$ ...
\end{promptbox}

\noindent\textit{Sample Prompt for NMI Analysis on TriviaQA:}

\begin{promptbox}
Question: How old was Jimi Hendrix when he died?

Answer: 27

\vspace{6pt}

Question: Who was the next British Prime Minister after Arthur Balfour?

Answer: henry campbell bannerman

\vspace{6pt}

Question: In what year's Olympics were electric timing devices and a public-address system used for the first time?

Answer: in 1912 in stockholm

\vspace{6pt}

Question: Where did the Shinning Path terrorists operate?

Answer: republic of perú

\vspace{6pt}

Question: What was the last US state to reintroduce alcohol after prohibition?

Answer: history of mining in utah ...

\vspace{6pt}

\end{promptbox}

\subsection{Results on Qwen Models of Various Sizes}
\label{qwen_twins}
\begin{figure}[H]
    \centering
    \includegraphics[width=0.9\textwidth]{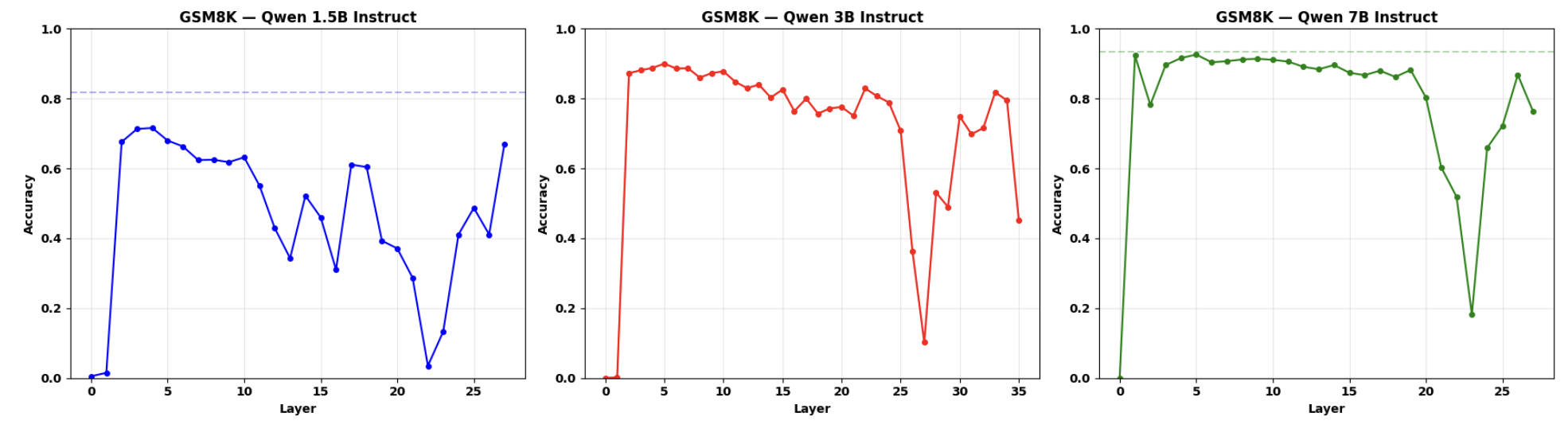}
    \caption{Zero ablation on Qwen2.5-(1.5B,3B,7B)-Instruct Models. We find that all 3 models have critical layers at relatively similar positions.}
    \label{fig:methodology}
\end{figure}

\subsection{Testing Logical and Airthmetic Errors} 

\begin{figure}[H]
    \centering
    \includegraphics[width=1.0\textwidth]{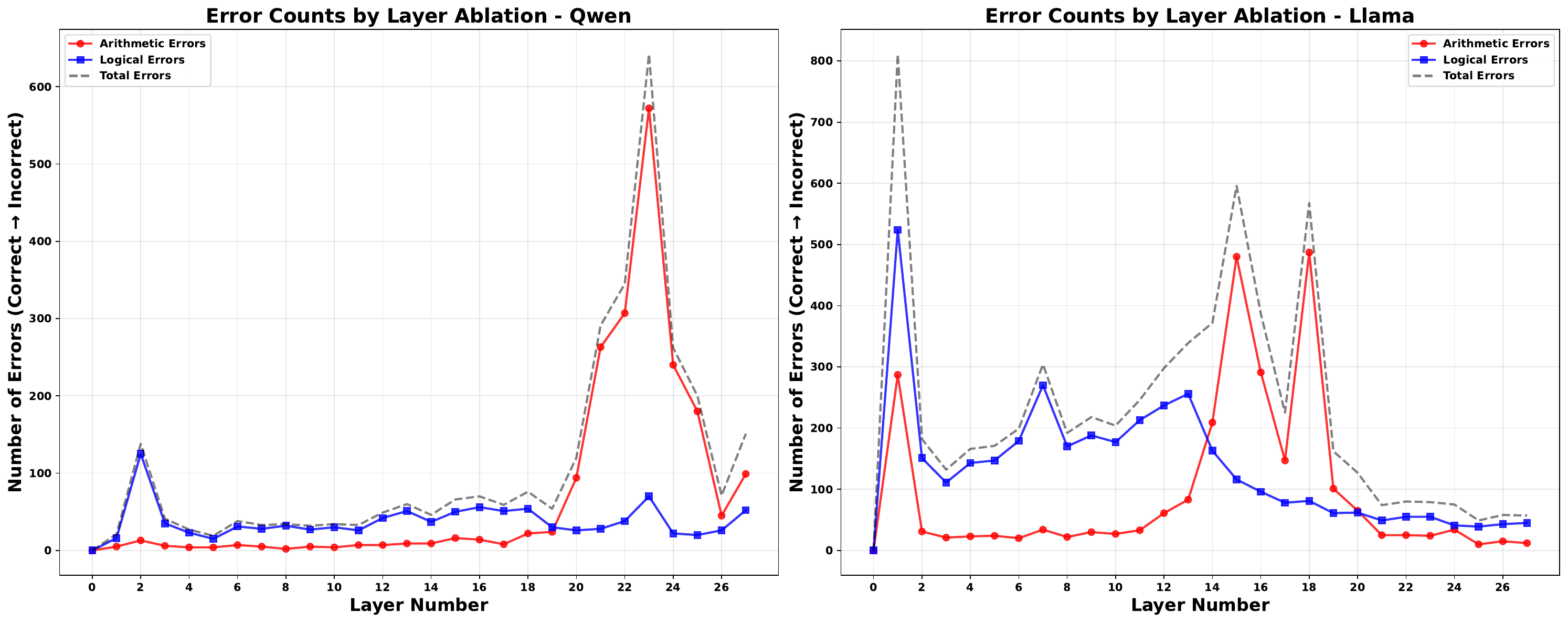}
    \caption{Error counts by layer ablation for Qwen (left) and Llama (right) instruct models. We used the GPT API to automatically classify model responses into arithmetic errors (red) and logical errors (blue). Notably, for Qwen, layer 23 shows a strong spike in arithmetic errors despite logically correct reasoning, suggesting that arithmetic mistakes dominate at critical layers. In contrast, logical errors remain more evenly distributed across layers. For the Qwen base model (not shown here), early layers (e.g., layer 2) produced repetitive and incoherent responses} 
    \label{fig:arith_logic}
\end{figure}

\subsection{TriviaQA NMI Plot} 

\begin{figure}[H]
    \centering
    \includegraphics[width=1.0\textwidth]{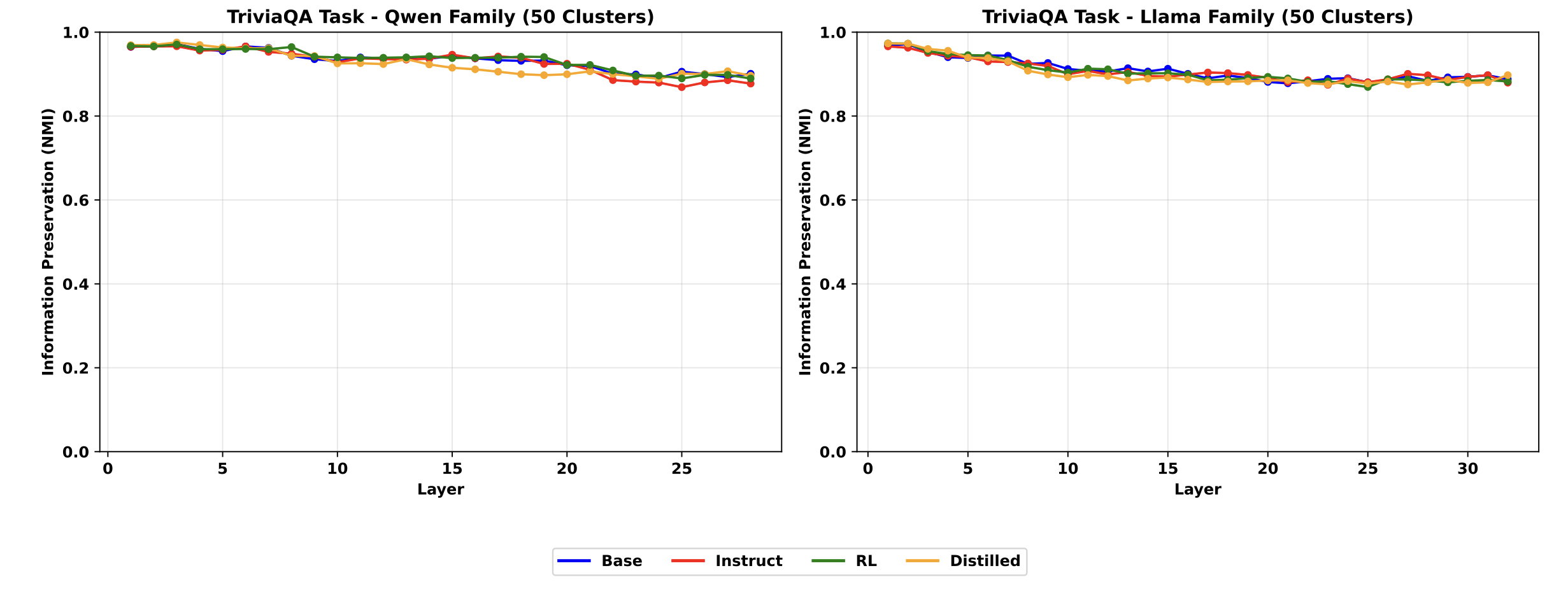}
    \caption{NMI remains relatively stable for triviaQA} 
    \label{fig:trivia_nmi}
\end{figure}

\subsection{Residual Norm analysis}
\label{residual}

\begin{figure}[H]
    \centering
    \includegraphics[width=1.0\textwidth]{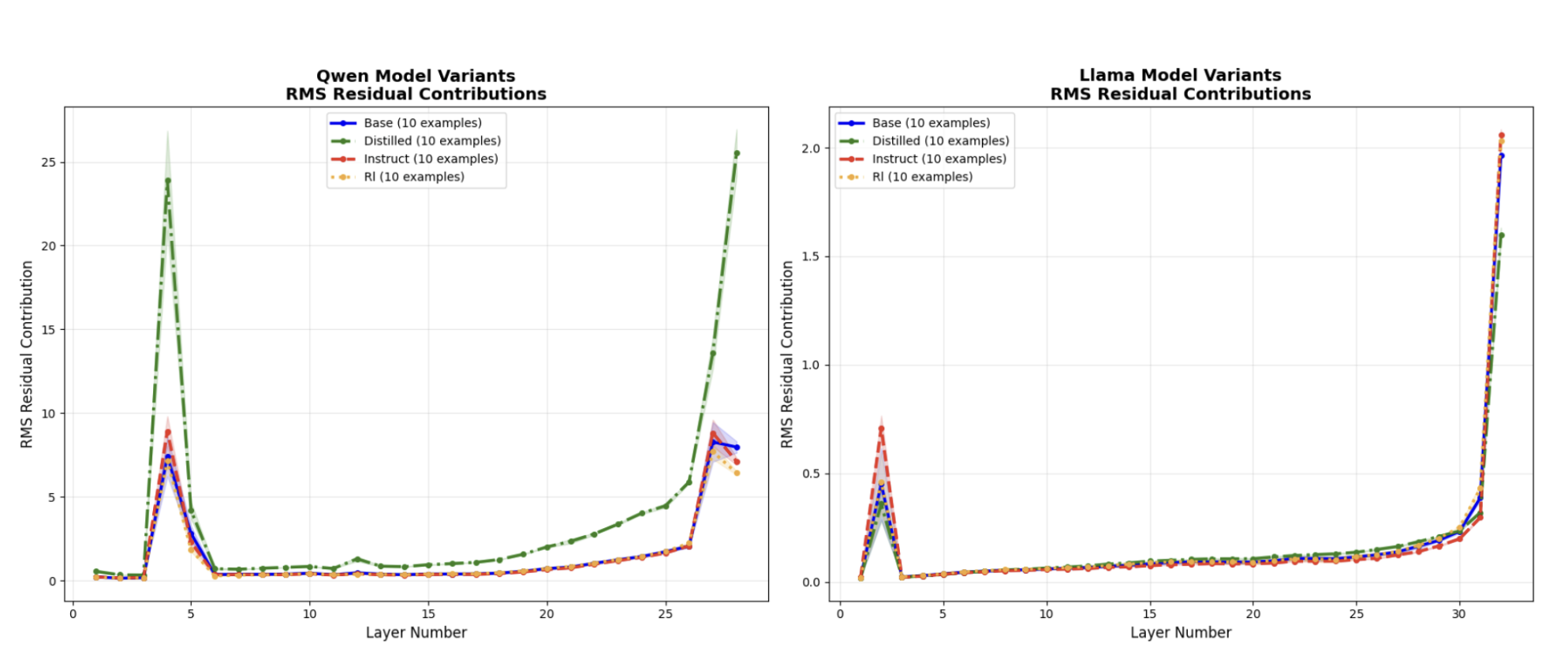}
    \caption{Root-mean-square (RMS) norm of residual stream contributions by layer for Qwen (left) and LLaMA (right) model variants. The curves show how much each layer adds to the residual stream, with largely consistent patterns across Base, Distilled, Instruct, and RL variants.}
    \label{fig:rms_norm}
\end{figure}

\end{document}